\documentclass[runningheads]{llncs}
\usepackage[T1]{fontenc}
\usepackage{graphicx,verbatim}
\usepackage{multirow}
\usepackage{booktabs}
\usepackage{color}
\usepackage{url}
\usepackage[hidelinks]{hyperref}
\usepackage{xcolor}
\hypersetup{
    colorlinks=true, 
    urlcolor=blue,  
    linkcolor=blue, 
    citecolor=blue   
}

\usepackage{enumitem}
\usepackage{amssymb}
\usepackage{amsmath}
\begin{document}
\title{Anatomy-Aware Prediction of Bronchoscopic Accessibility from 3D CT}
%

\author{Linkai Peng\inst{1} \and Cuiling Sun\inst{2} \and Bin Wang\inst{1} \and Jamie Rowell\inst{3} \and Catherine Gao\inst{3} \and Oyku Ikizgul\inst{4} \and Eminenur Sen Tasci\inst{5} \and Andrea Bejar\inst{5} \and Halil Ertugrul Aktas\inst{5} \and Gorkem Durak\inst{5} \and Momen Wahidi\inst{3} \and Christopher Kapp\inst{3} \and Ulas Bagci\inst{5}\dag}
\index{Peng, Linkai}
\index{Sun, Cuiling}
\index{Wang, Bin}
\index{Rowell, Jamie}
\index{Gao, Catherine}
\index{Ikizgul, Oyku}
\index{Tasci, Eminenur Sen}
\index{Bejar, Andrea}
\index{Aktas, Halil Ertugrul}
\index{Durak, Gorkem}
\index{Wahidi, Momen}
\index{Kapp, Christopher}
\index{Bagci, Ulas}
\authorrunning{L. Peng et al.}
\institute{Department of Electrical and Computer Engineering, Northwestern University \and Department of Computer Science, Northwestern University \and Department of Medicine, Northwestern University \and Department of Radiology, Istanbul Faculty of Medicine \and Department of Radiology, Northwestern University \\
$\dagger$~Corresponding author: \email{ulas.bagci@northwestern.edu}
    }
  
\maketitle              
\begin{abstract}
Pre-operative planning for bronchoscopy is critical for the diagnosis of lung lesions. Current accessibility assessment relies on subjective manual inspection of CT scans, which is time-consuming and prone to inter-observer variability. In this paper, we formalize bronchoscopy accessibility prediction as a novel supervised learning task and present the first end-to-end framework to address it. We propose an \textbf{Anatomy-Aware Mixture-of-Experts (MoE)} model that integrates specialized modules: a CT Expert for local morphological features, a Lobe Expert for anatomical priors, and a Path Geometry Expert that encodes the sequential constraints of the bronchial tree. To support this task, we curated \textbf{the first clinical dataset of 438 cases} with pre-operative CT scans and documented procedural outcomes. Experimental results demonstrate that our method achieves an AUROC of 0.8052, significantly outperforming both state-of-the-art baselines and experienced human experts. This work establishes a new benchmark for computer-aided interventional planning in pulmonary medicine. Our data and code will be publicly available at \url{https://nubagcilab.github.io/BronchoAccess/}.


\end{abstract}
\section{Introduction}

Bronchoscopy is widely used for the diagnosis of pulmonary lesions. 
Although modern navigation systems assist intra-procedural guidance \cite{folch2019electromagnetic,khandhar2017electromagnetic}, pre-operative accessibility assessment remains largely subjective. As demonstrated in our benchmark, expert bronchoscopists predict a diagnostic procedure with an AUROC of only 0.5661. This performance highlights the intrinsic difficulty of the task.

Despite its clinical relevance, bronchoscopy accessibility prediction remains underexplored. Prior work has focused on airway segmentation \cite{wang2022naviairway,yang2024airway,zhang2023multi}, lesion detection\cite{nasrullah2019automated}, or path planning \cite{ciobirca2018new}. While these tasks provide an important basis, they do not directly address procedural accessibility. Existing attempts at outcome prediction rely on handcrafted geometric descriptors and statistical association models without end-to-end supervision \cite{naito2023predicting}, failing to capture nonlinear interactions between local morphology, anatomical location, and airway complexity. A standardized benchmark for this task is also absent.

To bridge this gap, we first formalize bronchoscopy accessibility prediction as a supervised learning task and curate the first dataset dedicated to this problem. We retrospectively collected 438 clinically verified cases from three medical centers, each with documented procedural outcomes following bronchoscopy for lung lesions. This multi-center cohort provides diverse anatomical variability and realistic clinical distributions.

Building upon this task definition, we propose an anatomy-aware mixture-of-experts (MoE) \cite{shazeer2017outrageously} framework that decomposes accessibility prediction into three components. The CT Expert captures local volumetric context around the lesion. The Lobe Expert encodes anatomical priors reflecting regional variations. The Path Geometry Expert models cumulative geometric constraints along the airway trajectory. These representations are adaptively fused through a gating mechanism that performs case-dependent expert weighting.

To enable systematic evaluation, we further establish the first comprehensive benchmark for bronchoscopy accessibility prediction. The benchmark includes traditional machine learning methods, CT-based deep networks, graph neural networks, and human expert assessment. This benchmark provides a unified evaluation framework for future research on data-driven procedural planning.

Our contributions are three-fold:
\begin{enumerate}[nosep, font=\bfseries]
    \item To address this severely under-studied clinical challenge, we formalize bronchoscopy accessibility prediction as an end-to-end supervised deep learning task. To the best of our knowledge, this work presents the first deep learning application dedicated to predicting functional procedural outcomes directly from pre-operative CT scans.
    \item We introduce an anatomy-aware MoE architecture that integrates volumetric imaging, anatomical priors, and airway geometry within a unified model.
    \item We establish the first comprehensive benchmark for bronchoscopy accessibility prediction using 438 clinically verified cases and evaluate our framework against diverse computational baselines and human expert assessments.
\end{enumerate}

\section{Method}

\begin{figure}[!htbp]
\includegraphics[width=\textwidth]{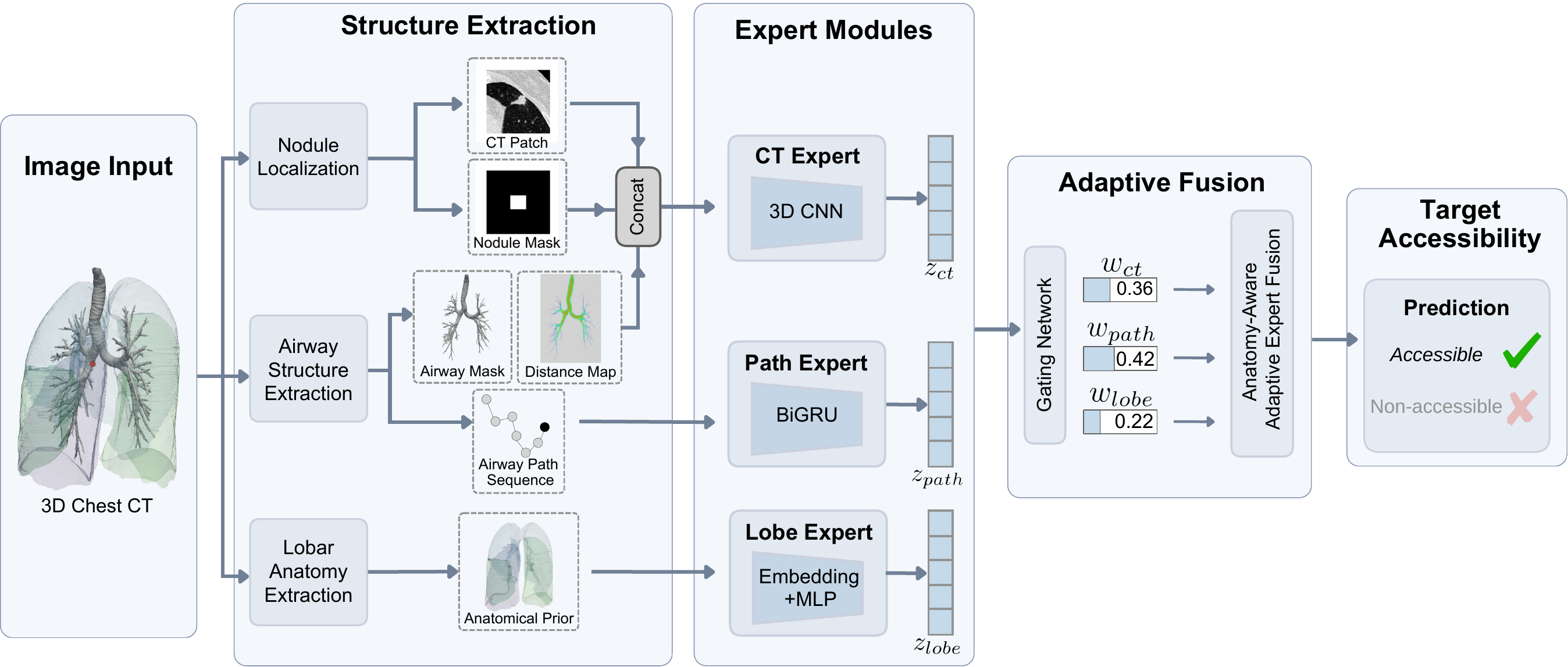}
\caption{Overview of the proposed anatomy-aware mixture-of-experts framework for bronchoscopy accessibility prediction.} \label{main}
\end{figure}

Given CT volume $X \in \mathbb{R}^{H \times W \times D}$, airway tree graph $\mathcal{G} = (\mathcal{V}, \mathcal{E})$, target lesion centroid $\mathbf{c}_N \in \mathbb{R}^3$ and lobe identifier $L \in \{1, \dots, 5\}$, we learn $\mathcal{F}: \{X, \mathcal{G}, \mathbf{c}_N, L\} \rightarrow y$, where $y \in \{0, 1\}$ denotes bronchoscopy accessibility. Our framework decomposes this into three components as shown in Fig. \ref{main}.

\subsection{Expert Modules}
\subsubsection{CT Expert}
The local structural relationship between the target lesion and the surrounding airway tree plays a central role in bronchoscopy accessibility. 
CT Expert is formulated to capture the local morphological and spatial relationship between the target lesion and its surrounding airway structures.

We extract a multi-channel 3D patch $x$ comprising the raw CT intensities, a binary lesion mask, and a 3D Euclidean Distance Transform (EDT) \cite{meijster2000general} of the airway. The lesions are first automatically detected and subsequently verified to ensure consistency with the procedural target. The EDT provides continuous spatial gradients that stabilize learning compared to sparse airway masks. In this way, the model receives an explicit geometric prior reflecting airway accessibility rather than relying solely on implicit learning from intensity patterns.

The multi-channel patch is processed by a custom 3D convolutional encoder. It consists of 4 consecutive residual blocks \cite{he2016deep} integrated with squeeze-and-excitation (SE) \cite{hu2018squeeze} channel attention. After convolution, a global average pooling layer aggregates the spatial features and a linear layer projects it into a latent embedding $z_{ct} \in \mathbb{R}^{d}$, where $d$ is the joint feature space dimension.

\subsubsection{Lobe Expert}
Bronchoscopy difficulty varies across pulmonary lobes due to anatomical differences in branching depth and orientation. To incorporate this anatomical prior, we encode the lobe identity $L$ as a categorical variable and map it into a learnable embedding space via an embedding layer followed by a two-layer MLP: $
z_{lobe} = \text{MLP}(\Psi(L)), z_{lobe} \in \mathbb{R}^d.
$
This embedding serves as a global anatomical prior that regularizes the final decision. By encoding lobe identity as a continuous representation, the model can learn structured similarities between lobes rather than treating them as independent categories.

\subsubsection{Path Geometry Expert}
Accessibility is constrained by cumulative geometric resistance along the airway trajectory from the trachea to the target region. These constraints are sequential and longitudinal in nature, which motivates explicit modeling of the airway trajectory. While the airway tree is naturally a graph, once the navigation path is selected, the data along it forms a one-dimensional ordered sequence of geometric descriptors; we therefore adopt a sequence model rather than a graph neural network. We represent the airway tree as a graph $\mathcal{G}$ derived from the segmented airway and extract the candidate navigation path toward the lesion. For cases with multiple plausible routes, the path with the minimal Euclidean distance to the lesion centroid is selected. Nodes along this path are ordered according to anatomical generation to form a proximal-to-distal sequence. To ensure robustness to segmentation noise, we retain only the largest connected component and remove edges shorter than a predefined physical threshold.

At each node, we compute a set of geometric and topological descriptors, including airway diameter, branching index, curvature, torsion, and relative path length. The ordered feature sequence forms a representation $S = \{\mathbf{s}_1, \dots, \mathbf{s}_T\}$, where $T$ differs across patients depending on airway depth.

To model longitudinal constraints, we apply a 1-layer bidirectional Gated Recurrent Unit (GRU) \cite{cho2014learning} over the sequence. The GRU is configured with a hidden size of 128 per direction. The final hidden states from both directions are concatenated and then projected to the joint feature space to obtain a path representation $z_{path}$. By explicitly modeling the airway as a sequence, this expert captures progressive narrowing, branching accumulation, and distal complexity. This design provides the framework with a topological understanding of accessibility that is complementary to the visual patterns captured by the CT Expert.

\subsection{Anatomy-Aware Adaptive Expert Fusion}
To handle cross-case heterogeneity, we employ an adaptive mixture-of-experts fusion. A gating network assigns importance weights $\mathbf{w} = [w_{ct}, w_{lobe}, w_{path}]^\top$ to calculate the final fused representation $Z = \sum w_i z_i$. The gating network conditions on case-level anatomical context: target lesion position $p$, lobe embedding $\Psi(L)$, and graph-level airway statistics $g$, producing weights via $w = \mathrm{Softmax}(MLP([p; \Psi(L); g]))$. It is implemented as a three-layer MLP with ReLU activations and a final Softmax layer. The fused representation $Z$ is then passed through a classification head to produce the probability of bronchoscopy accessibility $\hat{y} \in [0, 1]$. The entire system is trained using the binary cross-entropy (BCE) loss $\mathcal{L}_{bce}$. To encourage balanced expert utilization, we add an entropy regularization term on the gating weights:
\begin{equation}
    \mathcal{L}_{ent} = -\frac{1}{M} \sum_{i=1}^{M} \sum_{j \in \{ct, lobe, path\}} w_{i,j} \log(w_{i,j} + \epsilon),
\end{equation}
where $M$ denotes the number of samples and $\epsilon$ is a small constant for numerical stability. We encourage higher entropy in the gating distribution to avoid degenerate solutions where a single expert dominates across all cases. The total training loss $\mathcal{L}_{total}$ is formulated as:
$\mathcal{L}_{total} = \mathcal{L}_{bce} - \lambda \mathcal{L}_{ent},$
where $\lambda$ is a hyperparameter controlling the regularization strength. 

\section{Experiments}
\subsection{Dataset}
To fully evaluate the proposed framework, we curated a retrospective multi-center cohort consisting of 438 patients who underwent bronchoscopy for lung lesions. The dataset comprises cases from three clinical centers, including 279 successful procedures and 159 failed attempts. A procedure is labeled successful when diagnostic tissue is confirmed by histopathological analysis of the biopsy sample, which is an objective outcome based on diagnostic yield. Accessibility labels are derived from documented clinical outcomes. All procedures were performed using the same platform to ensure consistency in primary navigation technology across cases. To our knowledge, this is currently the largest dataset reported for outcome-driven bronchoscopy accessibility prediction. The dataset also involves multiple operators and varying device configurations. Although these procedural factors were not explicitly modeled, the multi-center design introduces heterogeneity that reduces systematic bias from a single operator or device type. To ensure robust evaluation, we employ 5-fold cross-validation with patient-level splits.

\subsection{Implementations and Compared Methods}
For the CT Expert, 3D patches of $128 \times 128 \times 32$ are extracted around the target lesion. Intensity values were clipped to the lung window range and normalized. We used RetinaNet \cite{baumgartner2021nndetection,cardoso2022monai,lin2017focal} for lung lesion detection and Naviairway \cite{wang2022naviairway} for airway segmentation. All models are trained using AdamW with a learning rate of $1 \times 10^{-3}$, batch size 16, and 100 epochs on an NVIDIA A100 GPU. To provide clinical context, 22 clinicians independently assessed accessibility from CT volumes and target locations. Three-level difficulty ratings were mapped to probabilistic scores (0.9, 0.7, 0.4) to compute AUROC and PRAUC. The coefficient $\lambda$ is set to 0.01. Dropout and weight decay were used during training to prevent overfitting.
We compare against conventional machine learning models trained on handcrafted geometric descriptors, CT-based deep networks (ResNet-18 \cite{he2016deep}, DenseNet-121 \cite{huang2017densely}, EfficientNet-b0 \cite{tan2019efficientnet}, ConvNeXt \cite{liu2022convnet}, and Swin Transformer \cite{liu2021swin}), and graph neural networks (GCN \cite{kipf2016semi}, GAT \cite{velivckovic2017graph}, GIN \cite{xu2018powerful}, and GraphSAGE \cite{hamilton2017inductive}) operating on airway topology.

\section{Results}

\subsection{Quantitative Results}
\subsubsection{Comparison with Diverse Baselines}
Table \ref{mainresults} presents the overall comparison across various methods. Conventional machine learning models trained solely on handcrafted geometric features show limited discriminative ability. CT-based deep networks achieve moderate performance, with the best AUROC reaching 0.6333. Graph neural networks slightly improve structural modeling, achieving up to 0.6483 AUROC. Human experts achieve an AUROC of 0.5661, indicating that visual CT inspection alone is insufficient for reliable accessibility estimation. 

The proposed anatomy-aware mixture-of-experts model achieves an AUROC of 0.8052 and PR-AUC of 0.8496, significantly outperforming computational baselines and human assessment. These results demonstrate the benefit of explicitly integrating airway trajectory modeling with local imaging cues. Importantly, the proposed model maintains competitive computational efficiency. With 3.83M parameters, it is more lightweight than most CT-based backbone networks while achieving superior predictive performance. This efficiency supports potential integration into pre-procedural planning workflows.
\begin{table}[!htbp]
\caption{Performance comparison of the proposed method against various baseline models. Human assessments were provided by 22 clinicians based on CT volumes and target lesion locations. $^\dagger$ indicates statistically significant difference compared with the proposed method (DeLong test, p < 0.05).}
\label{mainresults}
\setlength{\tabcolsep}{1pt}
\centering
\begin{tabular}{c|l|cccc|cc}
\hline
Input & Method & Acc. & F1 & AUROC & PRAUC & Params\\
\hline
\multirow{2}{*}{\shortstack{Geometric\\Features}} & Logistic Regression\cite{hosmer2013applied}& 0.6136 & 0.7536 & 0.5128 & 0.6897 & N/A \\
& Random Forest\cite{breiman2001random} & 0.5682 & 0.7031 & 0.4590 & 0.5785 & N/A\\
\hline
\multirow{5}{*}{\shortstack{CT\\Volumes}} & Densenet121\cite{huang2017densely}$^\dagger$ & 0.5909 & 0.7313 & 0.5301 & 0.6543 & 11.24M\\
& SwinTransformer\cite{liu2021swin}$^\dagger$ & 0.6136 & 0.7499 & 0.5418 & 0.8069 & 38.50M\\
& Efficientnetb0\cite{tan2019efficientnet}$^\dagger$ & 0.6022 & 0.7517 & 0.5725 & 0.7459 & 4.69M \\
& ConvNeXt\cite{liu2022convnet}$^\dagger$ & 0.6023 & 0.7482 & 0.5982 & 0.7750 & 31.30M\\
& Resnet18\cite{he2016deep}$^\dagger$ & 0.5795 & 0.6185 & 0.6333 & 0.7700 & 33.16M\\
\hline
\multirow{4}{*}{\shortstack{Airway\\Graph}} & GAT\cite{velivckovic2017graph}$^\dagger$ & 0.6386 & 0.7794 & 0.5767 & 0.7115 & 52.35K \\
& GIN\cite{xu2018powerful}$^\dagger$ & 0.6265 & 0.7257 & 0.5774 & 0.6782 & 101.12K \\
& GraphSage\cite{hamilton2017inductive}$^\dagger$ & 0.6203 & 0.7541 & 0.5897 & 0.7129 & 86.14K \\
& GCN\cite{kipf2016semi}$^\dagger$ & 0.6329 & 0.7752 & 0.6483 & 0.7666 & 51.59K \\
\hline
Human Experts & N/A & 0.5272 & 0.5423 & 0.5661 & 0.7434 & N/A \\
\hline
{\bfseries Hybrid} & {\bfseries Ours} & {\bfseries 0.8068} & {\bfseries 0.8595} & {\bfseries 0.8052} & {\bfseries 0.8496} & 3.83 M  \\
\hline
\end{tabular}
\end{table}

\subsubsection{Comparison Under Identical CT Inputs}
To ensure that the performance gain does not arise from richer input representations alone, Table \ref{sameinput} compares our model with CT-based deep networks trained on identical multi-channel inputs, including CT intensity, airway distance maps, and lesion masks. Even under this controlled setting, our model maintains a clear advantage. The strongest CNN baseline achieves 0.7310 AUROC, while the proposed method reaches 0.8052 AUROC. This result indicates that architectural decomposition and adaptive expert fusion account for the performance improvement rather than input augmentation alone.

\subsubsection{Ablation Study}
Table \ref{ablation} evaluates the contribution of each expert and their combinations. Among single branches, the CT Expert achieves the strongest performance (AUROC 0.7382), confirming that local volumetric context provides a solid baseline. The Path Geometry Expert (0.6040 AUROC) captures meaningful structural constraints along the airway trajectory but remains insufficient alone. The Lobe Expert performs near chance level (0.5513 AUROC), consistent with its role as a global anatomical prior rather than a standalone predictor.
\begin{table}[!htbp]
\caption{Comparison with CT-based deep networks trained using multi-channel volumetric inputs (CT intensity, airway distance transform, and lesion mask). $^\dagger$ indicates statistically significant difference compared with the proposed method (DeLong test, p < 0.05).}
\label{sameinput}
\setlength{\tabcolsep}{4pt}
\centering
\begin{tabular}{l|cccc}
\hline
Method & Acc. & F1 & AUROC & PRAUC\\
\hline
SwinTransformer$^\dagger$ \cite{liu2021swin} & 0.6250 & 0.7626 & 0.5243 & 0.6698 \\
ConvNeXt$^\dagger$ \cite{liu2022convnet} & 0.6477 & 0.7438 & 0.6629 & 0.7057 \\
Densenet121$^\dagger$ \cite{huang2017densely} & 0.6590 & 0.7058 & 0.6640 & 0.7611 \\
Resnet18$^\dagger$ \cite{he2016deep} & 0.6818 & 0.7586 & 0.7109 & 0.7834 \\
Efficientnetb0 \cite{tan2019efficientnet} & 0.6704 & 0.7819 & 0.7310 & 0.8269 \\
\hline
{\bfseries Ours} & {\bfseries 0.8068} & {\bfseries 0.8595} & {\bfseries 0.8052} & {\bfseries 0.8496} \\
\hline
\end{tabular}
\end{table}
Pairwise combinations consistently improve over individual modules. Combining CT and Path Experts increases AUROC to 0.7756, demonstrating the complementary effect of trajectory modeling. The full model integrating all three experts achieves the best performance (0.8052 AUROC). We also observe that CT Expert receives higher weights for centrally located lesions, while Path Geometry Expert weights increase for distal and lower-lobe lesions, suggesting anatomically consistent specialization. The gating weights, $\mathbf{w} = [w_{ct}, w_{lobe}, w_{path}]^\top$, thus provide a built-in case-level attribution: for any prediction, $\mathbf{w}$ indicates which anatomical factor drives the decision, offering interpretability without requiring post-hoc explanation methods. These results indicate that bronchoscopy accessibility depends on the joint modeling of local morphology, anatomical context, and cumulative airway geometry rather than any single modality alone.

\begin{table}[!htbp]
\caption{Ablation study evaluating the contribution of each expert module. The full model integrates all experts through adaptive gating.}
\label{ablation}
\setlength{\tabcolsep}{4pt}
\centering
\begin{tabular}{ccc|cccc}
\hline
Lobe Expert & Path Expert & CT Expert & Acc. & F1 & AUROC & PRAUC\\
\hline
$\checkmark$ & - & - & 0.6190 & 0.7647 & 0.5513 & 0.7030 \\
- & $\checkmark$ & - & 0.6310 & 0.7704 & 0.6040 & 0.6665 \\
- & - & $\checkmark$ & 0.7500 & 0.8035 & 0.7382 & 0.8028 \\
\hline
$\checkmark$ & $\checkmark$ & - & 0.7045 & 0.7968 & 0.5960 & 0.6857 \\
$\checkmark$ & - & $\checkmark$ & 0.7841 & 0.8504 & 0.7477 & 0.8141 \\
- & $\checkmark$ & $\checkmark$ & 0.7500 & 0.8358 & 0.7756 & 0.8030 \\
\hline
$\checkmark$ & $\checkmark$ & $\checkmark$ & {\bfseries 0.8068} & {\bfseries 0.8595} & {\bfseries 0.8052} & {\bfseries 0.8496} \\
\hline
\end{tabular}
\end{table}

\subsection{Qualitative Analysis}
\begin{figure}[!htbp]
\includegraphics[width=\textwidth]{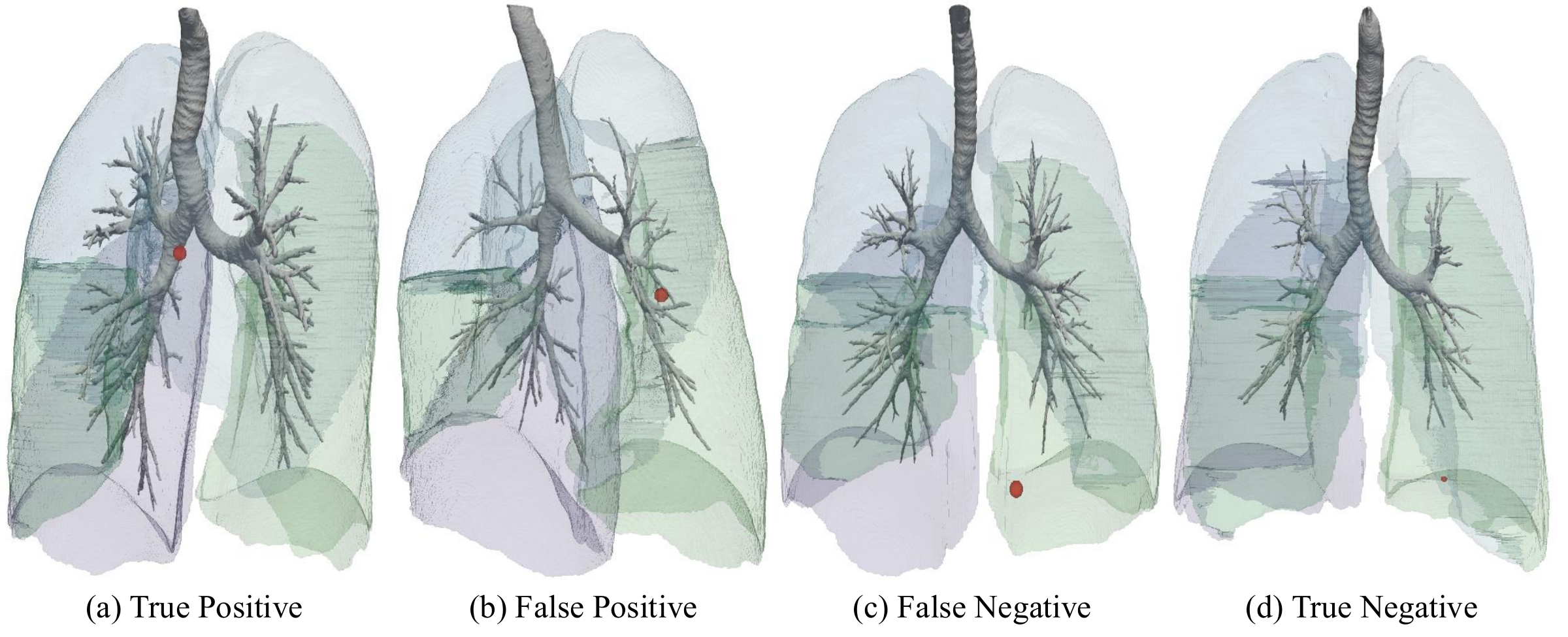}
\caption{Representative qualitative examples illustrating true positive (TP), false positive (FP), false negative (FN), and true negative (TN) predictions. Airway trees are visualized together with target lesion locations (red markers).} \label{visual}
\end{figure}

To further illustrate model behavior, Fig. \ref{visual} presents representative cases of true positive, false positive, false negative, and true negative predictions. The red marker indicates the target lesion location, and the airway tree is rendered to visualize the navigational trajectory.
The true positive case shows a lesion near a well-connected airway branch with sufficient diameter. The false positive case shows a lesion anatomically close to the distal bronchi but with abrupt branching changes along the path, where the CT Expert may overweight local proximity in such cases. The false negative involves a peripheral lesion with limited distal branches, where successful sampling may still occur in practice due to operator expertise or ultra-thin bronchoscopes. The true negative shows a deeply peripheral lesion with sparse connectivity. Overall, accessibility depends on both local lesion–airway adjacency and cumulative geometric constraints; clinical execution factors beyond static anatomy also influence outcomes.

\section{Conclusion}

To the best of our knowledge, this is the first study to systematically formalize bronchoscopy accessibility prediction as a supervised learning problem and to benchmark it against both computational baselines and human expert assessment. The results reveal that this clinically relevant task remains under-explored and difficult under conventional visual evaluation paradigms.

We presented the first learning-based framework for predicting bronchoscopy accessibility directly from pre-procedural CT. The proposed anatomy-aware mixture-of-experts model decomposes accessibility prediction into complementary components. It integrates local volumetric context, global anatomical priors, and sequential airway path geometry within a unified end-to-end architecture.

The experimental results demonstrate that our approach significantly outperforms both state-of-the-art single-modality baselines and experienced human experts. Notably, the model achieves these gains with a lightweight architecture, maintaining low parameter count and computational cost.

By bridging airway topology and functional procedural outcome prediction, this work moves beyond connectivity analysis toward quantitative accessibility assessment. We anticipate that such anatomy-aware modeling may support pre-procedural planning and improve patient selection in bronchoscopy-guided diagnosis.

\begin{credits}
\subsubsection{\ackname} This work was partially supported by NIH R01-HL171376.

\subsubsection{\discintname} The authors have no competing interests to declare that are relevant to the content of this article.
\end{credits}

%
%
%
\bibliographystyle{splncs04}
\bibliography{Paper-2633}

\end{document}